\documentclass[sn-mathphys-num]{sn-jnl}

\usepackage{graphicx}%
\usepackage{multirow}%
\usepackage{amsmath,amssymb,amsfonts}%
\usepackage{amsthm}%
\usepackage{mathrsfs}%
\usepackage[title]{appendix}%
\usepackage{xcolor}%
\usepackage{textcomp}%
\usepackage{manyfoot}%
\usepackage{booktabs}%
\usepackage{algorithm}%

\usepackage{algorithmicx}%
\usepackage{algpseudocode}%
\usepackage{listings}%
\usepackage{paralist}

\theoremstyle{thmstyleone}%
\theoremstyle{thmstyletwo}%

\theoremstyle{thmstylethree}%
\definecolor{dark_orange}{RGB}{198,95,16}

\begin{document}

\title[Article Title]{View Transformation Robustness for Multi-View 3D Object Reconstruction With Reconstruction Error-Guided View Selection}


\author{
    Qi Zhang\textsuperscript{\rm 1}, Zhouhang Luo\textsuperscript{\rm 2,1}, Tao Yu\textsuperscript{\rm 1}, Hui Huang\textsuperscript{\rm 1}\footnote{Corresponding author.}   \\  

    \textsuperscript{\rm 1}College of Computer Science and Software Engineering, \\
    Shenzhen University, Shenzhen, China\\
    \textsuperscript{\rm 2}Guangdong Laboratory of Artificial Intelligence and Digital Economy (Shenzhen), Shenzhen, China\\
    qi.zhang.opt@gmail.com, \{luozhouhang2022, yutao2023\}@email.szu.edu.cn, hhzhiyan@gmail.com
    


}









\maketitle

\begin{abstract}

 View transformation robustness (VTR) is critical for deep-learning-based multi-view 3D object reconstruction models, which indicates the methods' stability under inputs with various view transformations. However, existing research seldom focuses on view transformation robustness in multi-view 3D object reconstruction. One direct way to improve the models' VTR is to produce data with more view transformations and add them to model training. 
 Recent progress on large vision models, particularly Stable Diffusion models, has provided great potential for generating 3D models or synthesizing novel view images with only a single image input. 
 To fully utilize the power of Stable Diffusion models without causing extra inference computation burdens, we propose to generate novel views with Stable Diffusion models for better view transformation robustness. Instead of synthesizing random views, we propose a reconstruction error-guided view selection method, which considers the reconstruction errors' spatial distribution of the 3D predictions and chooses the views that could cover the reconstruction errors as much as possible.
 The methods are trained and tested on sets with large view transformations to validate the 3D reconstruction models' robustness to view transformations. Extensive experiments demonstrate that the proposed method can outperform state-of-the-art 3D reconstruction methods and other view transformation robustness comparison methods.
 Code is available at: \href{https://github.com/zqyq/VTR}{https://github.com/zqyq/VTR}.
 \end{abstract}

%

\section{Introduction}
Voxel-based multi-view 3D object reconstruction outputs 3D voxels of the object by fusing various viewpoints with deep-learning neural networks. It plays an important role in computer vision, robotics, augmented reality, and other domains. Naturally, multi-view 3D object reconstruction shall perform stably with inputs of different viewpoints, i.e., the view transformation robustness (VTR) is vital. However, few research works have studied the view transformation robustness for multi-view 3D object reconstruction.

\begin{figure}[t]
\begin{center}
   \includegraphics[width=0.8\linewidth]{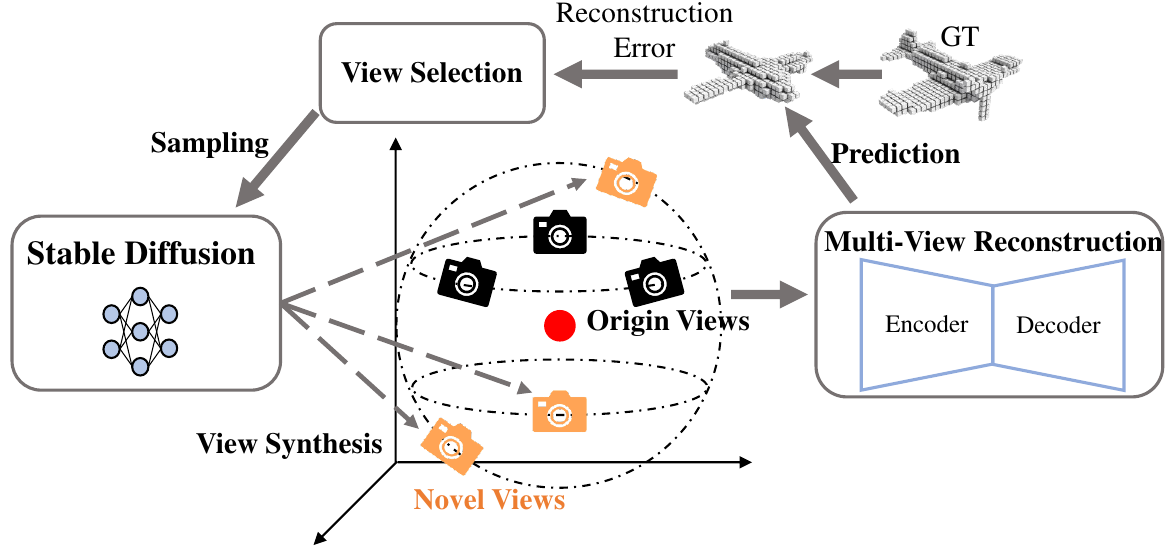}
\end{center}
   \caption{
   The method's main idea is to utilize the proposed reconstruction error-guided view selection method for selecting and generating viewpoints covering the most errors.}
\label{fig:general_idea}
\end{figure}

Instead, the VTR issue has been dealt with in other areas, such as in multi-view classification \cite{wei2020view} or point cloud recognition \cite{qi2017pointnet, Wang2022ARTPointIR}.
However, these methods are not specially designed to deal with the view transformation robustness issue for multi-view 3D object reconstruction. So, proposing novel methods for boosting view transformation robustness is important for multi-view 3D object reconstruction. 


Recent progress in large vision models (LVMs), especially stable diffusion models (SD) \cite{rombach2022high, hu2024neural, shi2023zero123++, qi2024vpp, yoo2024dreamsparse, zou2024sparse3d, kwak2024vivid, bauer2024challenges}, has provided the extra potential for multi-view 3D object reconstruction, such as training a large model for direct 3D object reconstruction \cite{liu2024one, liu2023syncdreamer, liu2023one2345++}, or novel view synthesis.
While utilizing LVMs for 3D reconstruction causes heavy computation burdens at the inference stage and their view transformation robustness \cite{Liu2023Zero1to3ZO} is still limited.
An effective way of utilizing stable diffusion models \cite{Liu2023Zero1to3ZO} for boosting the view transformation robustness of multi-view 3D reconstruction, is to use them as a data augmentation platform to aid the model training by randomly synthesizing novel views (see Figure \ref{fig:general_idea}).
However, even though these randomly generated views are useful for increasing the model's robustness to view transformations, the repeated or similar views are also being synthesized with no aid for the model's robustness, reducing the approach's effectiveness. 
Thus, \textbf{instead of generating random views, and considering the error spatial distributions of 3D reconstruction predictions, we propose a novel reconstruction error-guided view selection method to choose the most effective views to improve the 3D reconstruction models' view transformation robustness (VTR)}.

\begin{figure*}[t]
\begin{center}
   \includegraphics[width=\linewidth]{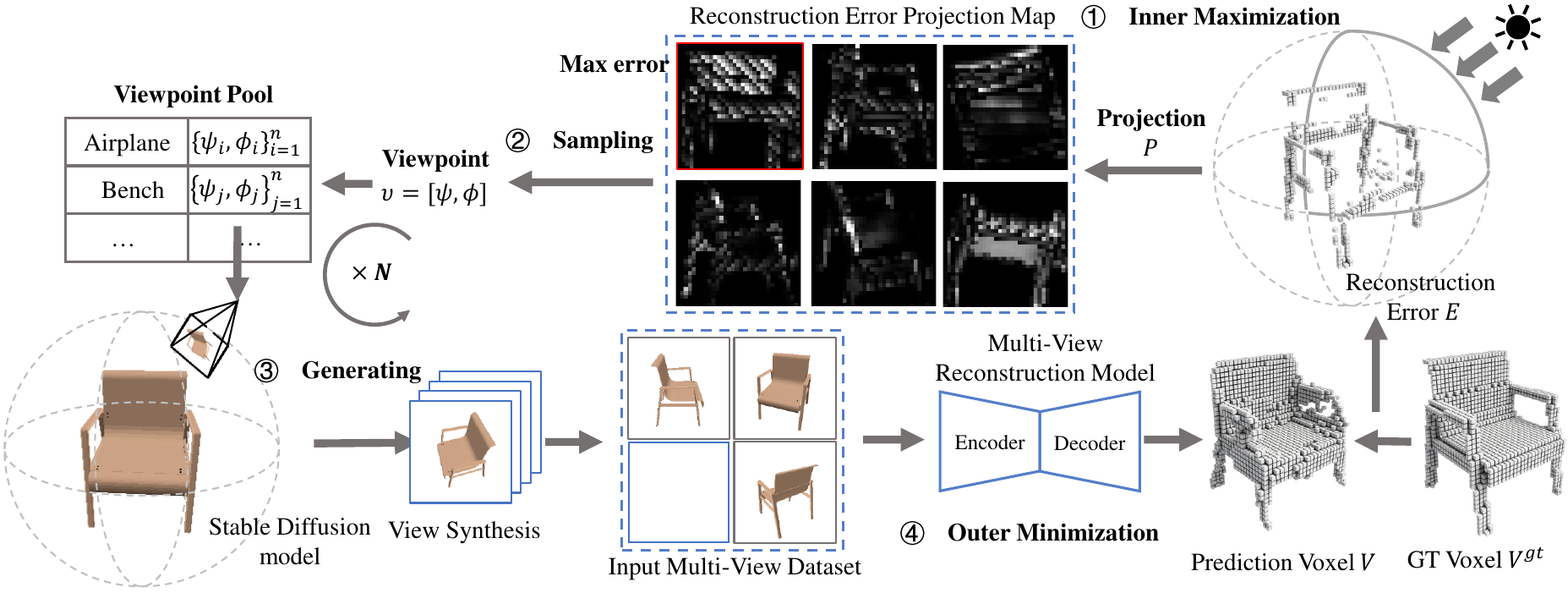}
\end{center}
   \caption{The pipeline of the method: first, the reconstruction error-guided view selection module chooses the views covering most reconstruction errors; then, the stable diffusion model works as a strong data platform for producing multi-view images of selected viewpoints; moreover, the new view images are added to finetune the 3D object reconstruction model.   
   These steps are alternatively conducted and trained to increase the 3D reconstruction models' robustness to view transformations.}
\label{fig:pipeline}
\end{figure*}

Overall, the proposed method consists of three components (see Figure \ref{fig:pipeline}): \textit{the original 3D object reconstruction model, the view selection module, and the stable diffusion model-based view synthesis module}. 
The original 3D object reconstruction model could be any deep learning-based multi-view 3D object reconstruction model (e.g., \citet{xie2020pix2vox++}, or \citealt{yang2023long}). It is first trained on the training set with limited view angles, thus it cannot perform well on the test set with large view transformations.
The view selection module is guided by the 3D reconstruction error, where the view angles which can cover more 3D object reconstruction errors are more likely to be selected. Then, the selected view parameters are fed into the view synthesis model to generate new views. 
The view synthesis module generates novel new views based on the stable diffusion model with the learned view parameters from the view selection module. Later, the newly synthesized views are added to the training set to fine-tune the 3D object reconstruction model further. 
These steps are recycled to gradually improve the 3D reconstruction model's view transformation robustness. 

Besides, since existing 3D object reconstruction datasets (e.g., ShapeNet \cite{chang2015shapenet}) are shot from view angels roughly around the objects with small viewpoint ranges (denoted as `Aligned' data), to study the view transformation robustness, we generate a new dataset \textit{ ShapeNet-VTR} with more view angle distributions: the `Hemispherical' and `Spherical' set 
(see Figure \ref{fig:dataset}). 
\textbf{The 3D reconstruction model is trained on the training set of `Aligned' data and tested on the `Hemispherical' and `Spherical' sets, to validate the method's robustness to view transformations} (see details in Sec. Experiments).
The paper's contributions are summarized as follows.
\begin{compactitem}
    \item This is the first study on view transformation robustness (VTR) for multi-view 3D object reconstruction. Seldom research has focused on the issue of view transformations in this area. Besides, we propose a new dataset specially designed for studying the issue. 
    \item We propose a novel reconstruction error-guided view selection method for choosing the most effective views, which is much more effective compared to randomly generating views or data augmentation methods.
    \item We use existing stable diffusion models for boosting the view transformation robustness of current 3D object reconstruction models, without training a new one or deploying them at the inference stage, without causing extra computation burdens. Thorough experiments demonstrate the advantages of the proposed method over 3D reconstruction SOTAs and other view transformation robustness comparison methods.
\end{compactitem}

\section{Related Work}

\subsection{Multi-view 3D object reconstruction}
Deep learning-based multi-view 3D object reconstruction methods have achieved remarkable performance.
Early methods focus on the fusion of multi-view features, such as \cite{su2015multi,paschalidou2018raynet,huang2018deepmvs,yao2018mvsnet}, where the multi-view features are reduced into a fixed size of feature maps. 
Later methods tried recurrent neural network fusion methods \cite{choy20163d,kar2017learning,dosovitskiy2020image,yao2019recurrent}, regarding input views as a sequence.
However, RNN fusion methods are not invariant to input order permutations and cannot handle a large number of views efficiently.
Attention-based fusion methods \cite{yang2020robust,xie2019pix2vox,zhu2023garnet,xie2020pix2vox++}
are also proposed to fuse multi-views according to the attention maps estimated from the attention subnet.
Recent methods utilize transformer networks for more complicated fusion between the views, such as \cite{shi20213d,tiong20223d,wang2021multi,yagubbayli2021legoformer,arshad2023list,yang2023long,zhu2023umiformer}. 
LRGT \cite{yang2023long} proposed long-range grouping attention for grouping tokens from all views with separate attention operations.
\textit{However, none of these methods have ever paid attention to the multi-view 3D object reconstruction methods' robustness to the input view transformations.}


\subsection{3D view transformation/rotation robustness}
The issue of view transformation and rotation robustness has been studied in various areas, such as point cloud classification and multi-view image classification.
Instead of using rotation invariant descriptors \cite{spezialetti2019learning,zhu2023r} as inputs or designing rotation equivariant networks \cite{shen20203d,cohen2021equivariant}, ART-Point \cite{Wang2022ARTPointIR} improved rotation robustness by training the point classifier on inputs with adversarial rotations.
Considering that the camera viewpoints are often fixed for all shapes in multi-view 3D shape classification, \citet{Hamdi2022MVTNLM} proposed the Multi-View Transformation Network (MVTN), which uses differentiable rendering to determine optimal view-points for 3D shape recognition. 
Recent methods have started to focus on novel view synthesis for boosting the model viewpoint robustness.
ViewFool \cite{Dong2022ViewFoolET} found adversarial viewpoints that mislead 3D recognition models with an entropic regularize. 
VIAT \cite{Ruan2023TowardsVV} proposed to improve the viewpoint robustness of multi-view image classification with the inner diverse adversarial viewpoints and the outer viewpoint invariant classifier training. 
\textit{Overall, synthesizing novel viewpoints and selecting the most optimal views are the keys for these methods to improve the models' view transformation or rotation robustness. Note that none of these methods have focused on the view transformation robustness issue for multi-view 3D object reconstruction.}

\section{Method}
The main idea of the method is to propose a novel reconstruction error-guided view selection method to select viewpoints that influence the reconstruction results most, then these viewpoint parameters are fed into an existing Stable Diffusion (SD) model to generate the selected view images for finetuning the original 3D reconstruction model for better view transformation robustness.
The proposed method consists of \textbf{the multi-view 3D reconstruction model}, \textbf{the view selection module}, and \textbf{the view synthesis module}, whose details are as follows.


\subsection{Multi-view 3D reconstruction model}

A multi-view 3D object reconstruction model $\mathit{f} $ usually consists of the following parts \cite{choy20163d}: First, an encoder to extract the feature representation for reconstruction from the image set ($I_1$,$\dots $,$I_n$) of Object $\mathit{I}$; Second, multiple 2D image features or 3D volume features are fed into a merger to fuse features from different views; Third, a decoder to predict the corresponding voxel-base 3D shape $\mathit{V} $ from the features map. 
The overall prediction process is formulated as: $V = f(I_1,\cdots ,I_n)$.
We adopt two typical kinds of multi-view 3D object reconstruction models, \textit{CNNs-based} Pix2Vox++ \cite{xie2020pix2vox++} and \textit{transformer-based} LRGT \cite{yang2023long}. 
Denote the 3D voxel ground truth as $V^{gt}$, and the reconstruction loss is $Loss(V, V^{gt})$ (details in Supp.).
Following \cite{xie2020pix2vox++} and \cite{yang2023long}, 
we use binary cross entropy loss in Pix2Vox++ while Dice loss \cite{milletari2016v} in LRGT. 
The model is first trained on the training set of `Aligned' data.



\begin{figure}[t]
\begin{center}
   \includegraphics[width=0.8\linewidth]{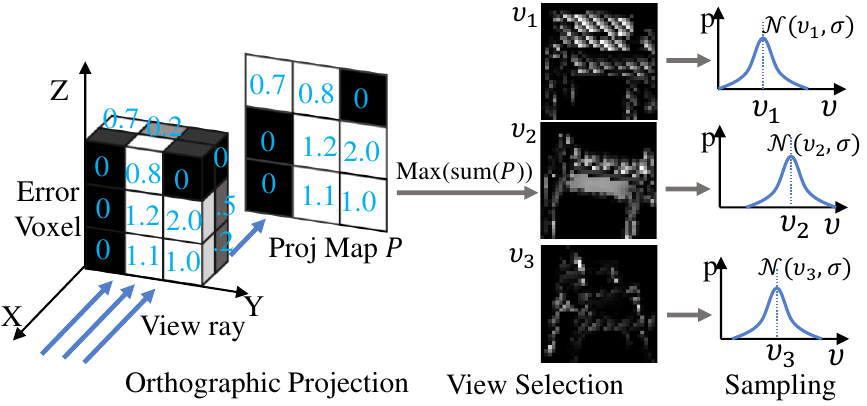}
\end{center}
\caption{The view selection module: select views covering most errors and then sample based on selected views.}
\label{fig:view_selection_module}
\end{figure}

\subsection{View selection module}

The view selection module first projects the 3D reconstruction error $\mathit{E}  = \left | V - V^{gt} \right |$ to a series of viewpoints, 
and then selects viewpoints containing the most reconstruction errors to be fed into the view synthesis module (see Figure. \ref{fig:view_selection_module}).  
Specifically,
We denote $v=[\psi, \theta, \phi]\in \mathbb{R} ^3$ as the viewpoint parameters, where $[\psi, \theta, \phi ]$ is the camera rotation along the z-y-x axes using the Tait-Bryan angles, and roll $\theta$ is set as zero. 
Then, our target is to estimate the $v=[\psi, \phi]$ from the 3D reconstruction error $\mathit{E}$,
which indicates the areas with bad reconstruction results of an object.
Naturally, if we could provide more information (views) about the areas with large reconstruction errors and try to rectify these errors, the performance of the reconstruction model could be boosted. 
Particularly, if a camera view can cover the reconstruction error regions more clearly, it is supposed to aid the robustness of the 3D reconstruction model.

To reduce computation, we first divide yaw $\psi$ and pitch $\phi$ into discrete values with $K$ degree interval to obtain a set of dense discrete 3D viewpoints. The yaw  is divided into $\left \{ \left [ -180^{\circ}\!+\!(i\!-\!1)\!\times K^{\circ}, -180^{\circ}+i\!\times K^{\circ} \right ]  \right \}_{i=1}^{n_\psi} $ and the pitch is divided into $\left \{ \left [ -90^{\circ}\!+\!(i\!-\!1)\!\times\! K^{\circ}, -90^{\circ}\!+\!i\!\times \! K^{\circ} \right ]  \right \}_{i=1}^{n_\phi} $, where $n_\psi=360^\circ/K^\circ$and $n_\phi=180^\circ/K^\circ$ refer the number of interval for the yaw $\psi$ and the pitch $\phi$, respectively. Then we traverse each interval and use the median value of each interval as the viewpoint position for that interval: $\psi=-180^{\circ}+(i-\frac{1}{2} )\times K^{\circ}$ and $\phi=-90^{\circ}+(i-\frac{1}{2} )\times K^{\circ}$.
As $K$ gets smaller, the traversed viewpoints $v=[\psi,\phi]$ are more continuous. 
By rotating the voxel grid $E$ by the corresponding degree $v$, we set the viewpoint direction to the negative direction along the X-axis, denoted as $E_{rot}=Rot(E, v)$, where $Rot$ is the rotation operation. 

We then perform the orthographic projection on $E_{rot}$ to each discrete viewpoint to obtain the reconstruction error projection maps $P$, conducted as follows:
\begin{equation}
    P((y,z),E_{rot})=E_{rot}(x^\ast, y, z),
\end{equation}
where the $x^\ast =\min_{E_{rot}(x ,y,z)\ne 0}X$, $E_{rot}(x,y,z)$ represents the integer 3D coordinate and the $X$ represents the range of $x$. Intuitively, we get the first occupancy value along each line of sight, ie., the surface of the error voxel. 

After we get the error projection map $P$ under each viewpoint $v$, we obtain the sum of the pixels of each projection map: $sum_{v}=\sum_{i=1}^{S}P_i $, where $S$ is the number of a projection map pixels and $P_i$ is the $i$-th pixel of the projection map $P$. Then, we select the maximum $n$ views $\{v_s\}$ based on the $sum_{v}$. Because larger $sum_{v}$ means the viewpoint covers more reconstruction error regions.
Rather than directly using the selected viewpoints for reconstruction, which will decrease the diversity of inputting viewpoints, we model each selected viewpoint $v_s$ as a Gaussian distribution and sample views among the distribution: $v\sim \mathcal{N} (v_s,\sigma )$, where $\sigma$ refers the standard deviation, which is set to $K/6$ in our experiments. Finally, we can obtain the $n$ corresponding viewpoints $v=[\psi, \phi]$ from the 3D reconstruction error $E$ for each object for $n$-view reconstruction. 


\textbf{Viewpoint pool.}
During experiments, we found that we construct a viewpoint set for each object under each category, the learned viewpoint distribution will degenerate with the increase of training iterations, that is, the diversity of selected viewpoints will decrease in the later training period, resulting in the phenomenon of overfitting in iterative training. A simple solution is to construct multiple viewpoint sets with multiple iterations for each object of each category. However, it will be very time-consuming. To solve this problem, according to \cite{Wang2022ARTPointIR}, it is observed that the selected viewpoint distributions of objects within the same category are highly similar, which has a strong transferability. Based on this, we construct a viewpoint pool by saving the viewpoints selected on each object by category:
\begin{equation}
    v=[\{v_{i,1} \}_{i=1}^{n_1},\dots ,\{v_{i,k} \}_{i=1}^{n_k},\dots ,\{v_{i,N} \}_{i=1}^{n_M} ],
\end{equation}
where $v_{i,k}$ refers to the viewpoint selected on object $i$ of category $k$. We will save the viewpoints corresponding to all $n_k$ objects in the category $k$ and traverse all $M$ categories to construct the viewpoint pool. In the process of iterative optimization, it is only necessary to sample the viewpoint pool according to the category and convert it into the corresponding view. Due to the transferability of the viewpoint distribution of the same type of object, high reconstruction loss can also be induced.

\begin{figure}[t]
\begin{center}
   \includegraphics[width=0.8\linewidth]{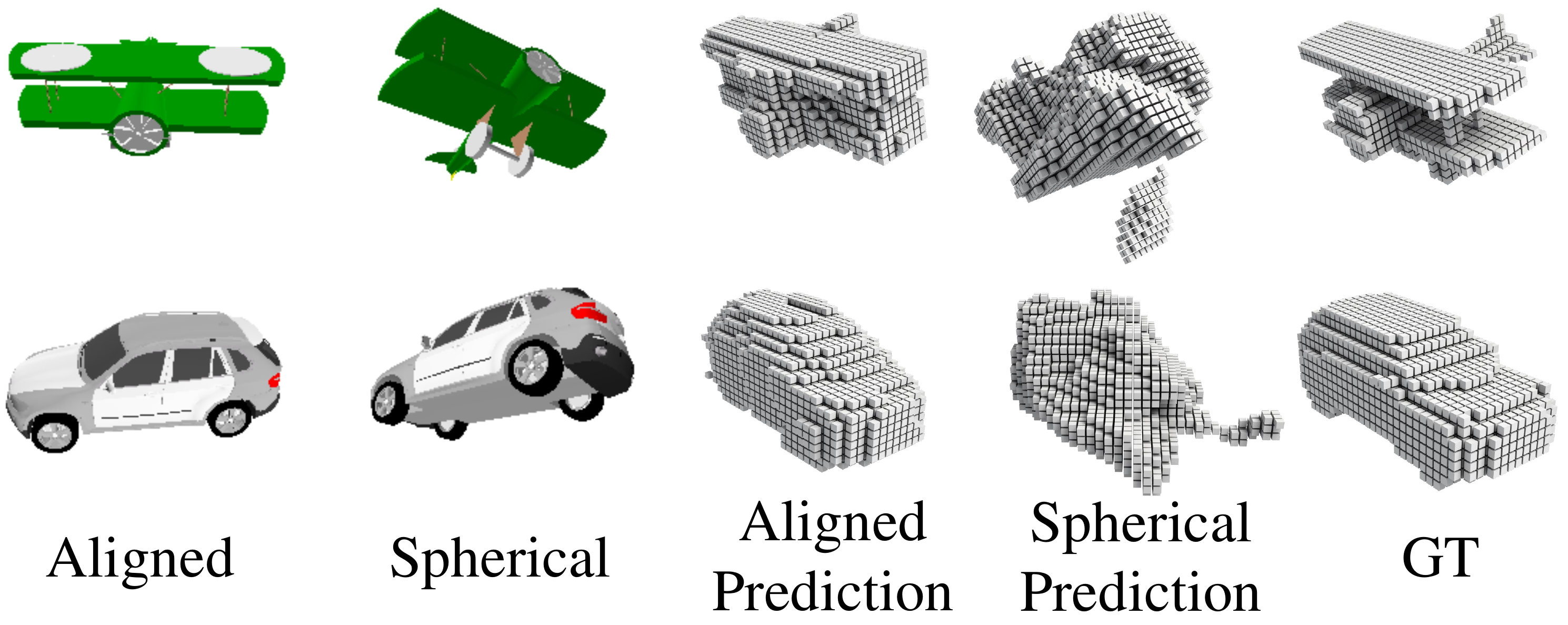}
\end{center}
\caption{The 3D reconstruction results of Zero-1-to-3, which has weak robustness to view transformations.}
\label{fig:stable_diffusion}
\end{figure}

\subsection{View synthesis from Stable Diffusion model}
With the camera viewpoint $v=[\psi, \phi]$ selected from the view selection module, we synthesize novel images of an object from the Stable Diffusion model Zero-1-to-3 \cite{Liu2023Zero1to3ZO}, which is a viewpoint-conditioned diffusion approach and can generate an image under the novel viewpoint by providing an image of the object and the viewpoint transformation matrix.
Given a dataset of paired images and their relative camera extrinsic {($\hat{I}$, $I$, $\hat{v}$, $v$)}, the model is trained by solving for the following objective:
\begin{equation}
    \min_{\theta }\mathbb{E}_{z\sim \varepsilon(\hat{I}),t,\epsilon \sim \mathcal{N}(0,1)} \parallel \epsilon -\epsilon _\theta (z_t, t, c(\hat{I}, I, \hat{v}, v))\parallel ^2_2,
\end{equation}
where $\varepsilon$ denote the encoder, $\epsilon _\theta$ is a denoiser U-Net, $t$ is the diffusion time step, and $z$ is the latent representation of the input image encoded by the encoder. After the model is trained, we can generate the novel image from any viewpoint by performing iterative denoising from a Gaussian noise image conditioned on $c(\hat{I}, \hat{v}, v)$.

\textbf{Use Zero-1-to-3 for direct 3D reconstruction.}
Directly using Zero-1-to-3 for 3D reconstruction reports bad performance as in Figure \ref{fig:stable_diffusion}.
It suffers drastically from the view transformations and the 3D reconstruction inference process is very time-consuming (see running time in the Supplemental). Thus, we utilize Zero-1-to-3 for novel view synthesis and rely on the new views to boost the view transformation robustness of existing 3D reconstruction models, instead of using it for direct 3D reconstruction.

\begin{figure}[t]
\begin{center}
   \includegraphics[width=0.8\linewidth]{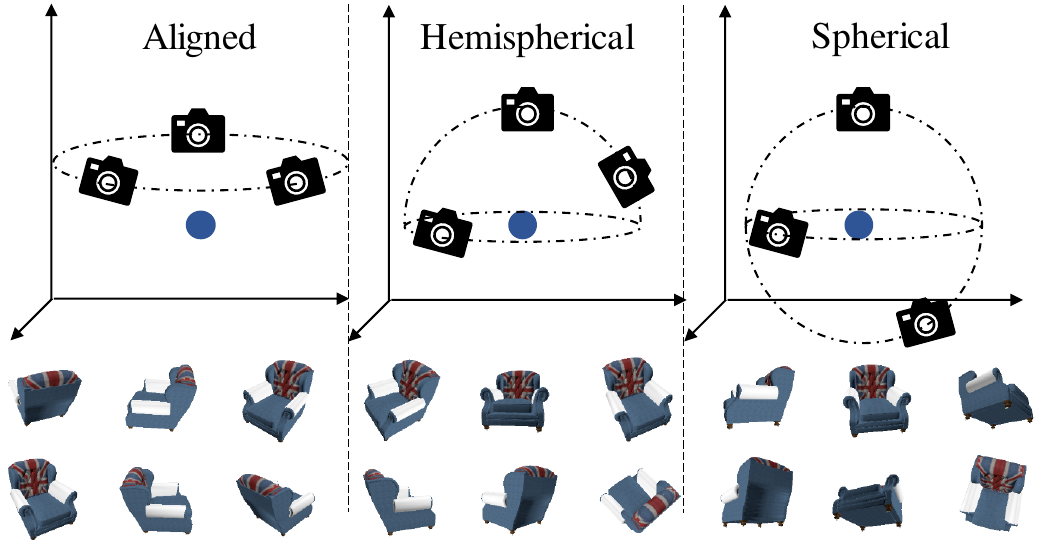}
\end{center}
\caption{The viewpoint distribution examples of our ShapeNet-VTR dataset. The blue dot is the object center.}
\label{fig:dataset} 
\end{figure}

\begin{table*}[t]
\small
\centering
\begin{tabular}
{l@{\hspace{0.12cm}}l@{\hspace{0.12cm}}|c@{\hspace{0.12cm}}c@{\hspace{0.12cm}}|c@{\hspace{0.12cm}}c@{\hspace{0.12cm}}|c@{\hspace{0.12cm}}c@{\hspace{0.12cm}}}
\hline
    &Test set &  \multicolumn{2}{c|}{Aligned}  &  \multicolumn{2}{c|}{Hemispherical}  &  \multicolumn{2}{c}{Spherical} \\
    &Method                                     & mIoU$\uparrow$    & F\_score$\uparrow$       & mIoU$\uparrow$    & F\_score$\uparrow$        & mIoU$\uparrow$    & F\_score$\uparrow$  \\
\hline
\multirow{2}{*}{None-VTR}
    & Pix2Vox++ \cite{xie2020pix2vox++}         & 64.63\%    & 42.02\%        & 47.02\%   &  26.49\%  & 36.71\%  & 18.44\%    \\
    & GARNet \cite{zhu2023garnet}               & 65.14\%    & 46.50\%        & 49.69\%   &  30.03\%  & 37.46\%  & 20.41\%    \\
\hline
\multirow{5}{*}{VTR}
    & Nature                                  & 65.54\%    & 46.69\%        & 49.00\%   &  30.20\%  & 37.13\%    & 21.03\%    \\
    & Random                                    & 64.89\%    & 46.67\%        & 50.89\%   &  31.77\%  & 41.62\%    & 24.45\%  \\
    & VIAT \cite{Ruan2023TowardsVV}            & 65.66\%    & \textbf{47.30\%}        & 51.50\%   &  31.88\%  & 40.36\%    & 23.47\%    \\
    & MVTN \cite{hamdi2022mvtn}                 & \textbf{65.67\%}    & 47.03\%        & 49.58\%   &  30.20\%  & 37.30\%    & 21.36\%    \\
    & Ours  (Pix2Vox++)                                    & 65.52\%    & 47.22\%        & \textbf{53.53\%}   &  \textbf{33.53\%} & \textbf{46.12\%}    & \textbf{27.01\%}     \\
\hline
\end{tabular}
\caption{The view transformation robustness comparison of CNNs-based 3D reconstruction methods.
None-VTR/VTR indicates methods without/with view transformation robustness techniques, respectively. All VTR use CNNs-based  Pix2Vox++ as the 3D reconstruction model. Our method achieves the best results on `Hemispherical' and `Spherical' with view transformations. }
\label{table:MV_results_CNNs}
\end{table*}

\begin{table*}[t]
\small
\centering
\begin{tabular}
{l@{\hspace{0.12cm}}l@{\hspace{0.12cm}}|c@{\hspace{0.12cm}}c@{\hspace{0.12cm}}|c@{\hspace{0.12cm}}c@{\hspace{0.12cm}}|c@{\hspace{0.12cm}}c@{\hspace{0.12cm}}}
\hline
    &Test set &  \multicolumn{2}{c|}{Aligned}  &  \multicolumn{2}{c|}{Hemispherical}  &  \multicolumn{2}{c}{Spherical} \\
    &Method                                     & mIoU$\uparrow$    & F\_score$\uparrow$       & mIoU$\uparrow$    & F\_score$\uparrow$        & mIoU$\uparrow$    & F\_score$\uparrow$  \\
\hline
\multirow{2}{*}{None-VTR}
    & UMIFormer \cite{zhu2023umiformer}   & 69.87\%    & 50.56\%  & 53.04\%     &  33.79\%  & 41.19\%  & 23.51\%    \\
    & LRGT \cite{yang2023long}            & 70.01\%    & 51.19\%    & 53.57\%   &  34.71\%  & 41.88\%  & 24.46\% \\
\hline
\multirow{5}{*}{VTR}
    & Nature  (LRGT)                               & 70.90\%    & 52.92\%        & 56.06\%   &  35.54\%  & 43.99\%    & 25.16\%    \\
    & Random (LRGT)                                & 70.92\%    & 52.68\%        & 56.02\%   &  35.28\%  & 44.18\%    & 25.32\%    \\

    & VIAT \cite{Ruan2023TowardsVV}  (LRGT)        & 70.86\%    & 52.33\%        & 56.07\%   &  35.29\%  & 44.70\%    & 25.27\%    \\
    & MVTN \cite{hamdi2022mvtn}  (LRGT)            & \textbf{71.09}\%    & \textbf{52.97}\%        & 55.76\%   &  34.92\%  & 43.58\%    & 24.07\%    \\
    & Ours (LRGT)                                  & 70.95\%    & 52.78\%        & \textbf{57.11\%}   & \textbf{35.89\%}   & \textbf{44.97\%}    & \textbf{26.02\%}     \\
\hline
\end{tabular}
\caption{The view transformation robustness comparison of Transformer-based 3D reconstruction methods.
 All VTR methods use Transformer-based LRGT as the 3D reconstruction model.
Our proposed method achieves the best results.}
\label{table:MV_results_transformer}
\end{table*}

\subsection{Training method}

The training adopts an iterative optimization scheme containing the inner maximization and the outer minimization.
Specifically, in the first iteration, we evaluate the pre-trained 3D reconstruction model to select the viewpoints for inner maximization, and then re-train the model on new view images generated by the view synthesis module from the previously selected viewpoints before obtaining a robust model for outer minimization. 
The process will be repeated until the model converges to the most robust state. Our goal is formulated as: $\min_{W}\sum_{i=1}^{N} \mathbb{E}_{v_i}  \left [ \max_{v_i}L(W,\mathcal{R}(v_i),y_i) \right ]$,
where $W$ denotes the parameters of the reconstruction model $f$, $L$ is a reconstruction loss function, $\mathcal{R}(v_i)$ is the rendered images of the $i$-th object given the viewpoint $v_i$ and the $y_i$ is the GT of the corresponding object. 
We then detail how we improve the inefficiency of the iterative optimization scheme with the random update strategy.

\textbf{Random update strategy.} 
A random update strategy is adopted to reduce the training time.
Instead of generating new views for all objects in each epoch, we randomly select 5\% objects and generate new views by Zero-1-to-3 for them, while the remaining 95\% objects are still trained with the views from the `Align' training set. At the same time, to make up for the diversity of viewpoints and improve the effectiveness of training, the newly generated view images will be added to the original view images set after each fine-tuning epoch, so that each epoch of iterative training can fully and effectively obtain the prior information left by the previous epoch, and help the 3D reconstruction model learn more efficiently.
\textbf{Note that only the multi-view 3D reconstruction model is needed at inference.}


\section{Experiments}

\subsection{Experiment settings}

\textbf{Dataset generation}.
Since no existing dataset is suitable for evaluating view transformation robustness, we generate a new dataset \textit{ShapeNet-VTR}, based on ShapeNet \cite{chang2015shapenet}, which consists of larger viewpoint ranges than the original version via rendering new views from the 3D CAD models in ShapeNet. ShapeNet-VTR consists of 13 categories and 39239 objects in total. The objects in each category are divided according to 7$:$1$:$2 for training, validation, and testing, respectively. ShapeNet-VTR consists of 3 sets: `Aligned', `Hemispherical', and `Spherical', sharing the same object dividing way but differing in the viewpoint range as shown in Figure \ref{fig:dataset}. 
`Aligned' set is rendered at a fixed $\phi$ of $60^{\circ}$ and every $15^{\circ}$ in $\psi$; In the `Hemispherical' set, the range of views is set as $\phi$ $\in$  $[0^{\circ}, 90^{\circ}]$, $\psi$ $\in$ $[-180^{\circ}, 180^{\circ}]$, and thus images are randomly rendered in the upper hemisphere. In `Spherical', the range of views is $\phi$ $\in$  $[-90^{\circ}, 90^{\circ}]$, $\psi$ $\in$ $[-180^{\circ}, 180^{\circ}]$,  and thus images are randomly rendered in the spherical space. In each set, each object contains 24 views but from different camera angle ranges. We train the 3D reconstruction models on the training set of  `Aligned' at first and \textbf{use the models' performance on `Hemispherical' and `Spherical' sets to evaluate the robustness of 3D object reconstruction models to view transformations}.

\textbf{Implementation details.}
We utilize two 3D reconstruction models, CNN-based Pix2Vox++ \cite{xie2020pix2vox++} and Transformer-based LRGT \cite{yang2023long}, respectively. 
The input view number is 3 and the image resolution is $256\times256\times 3$ and the size of the voxel output is $32\times32\times32$. We use a threshold of 0.4 to obtain the occupancy voxel grid and set the interval of degree $\mathit{K} = 30^{\circ} $ in the experiments.

\textbf{Comparison methods.} 
We compare with SOTA 3D object reconstruction methods Pix2Vox++ \cite{xie2020pix2vox++}, GARNet \cite{zhu2023garnet}, UMIFormer \cite{zhu2023umiformer}, and LRGT \cite{yang2023long} (without view transformation robustness approaches, denoted as None-VTR) and four view transformation robustness comparison methods (denoted as VTR). \textbf{Note that there are no existing VTR methods for multi-view 3D object reconstruction, and we propose reasonable baselines or adopt methods from other areas for comparison:}
\begin{compactitem}
    \item Nature: Data augmentation with the most common viewpoint renderings from training objects’ natural states (e.g. cars are usually viewed in the side view). 
    \item Random: Standard data augmentation with random viewpoints, and the view range is set as $\phi$ $\in$  $[-90^{\circ}, 90^{\circ}]$, $\psi$ $\in$ $[-180^{\circ}, 180^{\circ}]$.
    \item VIAT \cite{Ruan2023TowardsVV}: By regarding viewpoint transformation as an attack, VIAT solves the inner maximization problem to parameterize the Gaussian Mixture distribution of adversarial viewpoints with trainable parameters. In our experiment setting, VIAT keeps the roll $\theta$ fixed and learns the parameters of the Gaussian Mixture distribution of yaw $\psi$ and pitch $\phi$ to get an adversarial viewpoint. Next, we select the other two adversarial viewpoints at azimuth intervals of $120^\circ$ and $240^ \circ$, starting from the first adversarial viewpoint.
    \item MVTN \cite{hamdi2022mvtn}: A viewpoint selection algorithm that uses differentiable rendering to determine optimal viewpoints for 3D shape recognition through the gradient descent algorithm. We changed its downstream task to multi-view 3D reconstruction. We changed the 3D representation from point clouds to voxels and set the number of the selected viewpoints to 3 as the view amount of the input.
\end{compactitem}
Besides, we also evaluate the method's single-view reconstruction robustness, comparing with single-view reconstruction state-of-the-art methods \cite{mescheder2019occupancy}, which does not incorporate prior knowledge of point clouds or depth maps to assist in reconstruction as ours.

\textbf{Metrics.}
The evaluation metrics include the mean Intersection-over-Union (mIoU) and F\_score (see details in Supp.), whose higher values indicate better performance.

\subsection{Experiment results}

\subsubsection{Multi-view 3D reconstruction performance}
Table \ref{table:MV_results_CNNs} shows the view transformation robustness results of CNNs-based multi-view 3D object reconstruction methods. Our method utilizes Pix2Vox++ \cite{xie2020pix2vox++} as the 3D reconstruction model.
Compared with SOTA 3D reconstruction methods without view transformation robustness techniques (None-VTR), including Pix2Vox++ \cite{xie2020pix2vox++} and GARNet \cite{zhu2023garnet}, the proposed method outperforms them a lot on `Hemispherical' and `Spherical' with much larger view transformations than `Aligned'. 
\textbf{The proposed method also outperforms all VTR comparisons} (all use Pix2Vox++ as the 3D reconstruction model as ours) on both `Hemispherical' and `Spherical', indicating the proposed method is more robust to view transformation than all comparisons. `Nature' and `Random'  try to improve the model robustness via a data augmentation fashion, but also introduce redundant views to the model training, limiting the efficiency and effectiveness. VIAT \cite{Ruan2023TowardsVV} and  MVTN \cite{hamdi2022mvtn} are not specially designed for multi-view 3D reconstruction tasks: VIAT ignores the reconstruction error's spatial relation to the target views and MVTN relies on purely model learning for target view estimation without an in-depth rationale for the view selection step, and thus their performance is reduced.
In contrast, the proposed method considers the spatial connections between the 3D reconstruction error and the selected views, where the selected views could cover the reconstruction errors as much as possible to finetune the multi-view 3D reconstruction models on these views in the next circle. 

\begin{table}[t]
\small
\centering
\begin{tabular}{c|c|c|c}
\hline
    $K$ ($^{\circ}$) &  Aligned  &  {Hemispherical}  &  {Spherical} \\
\hline
    15                    & 65.37\%           & 51.44\% & 43.27\%    \\
    30                    &\textbf{65.52}\%  &\textbf{53.53}\%  &\textbf{46.12}\%   \\
    60                    &65.40\%  &52.72\%  &45.03\%  \\
    90                    &65.38\%  &52.32\%  &44.26\%    \\
\hline
\end{tabular}
\caption{The ablation study on the interval of degree $K$. $30^{\circ}$ achieves the best mIoU results. }
\label{table:ablation_angle}
\end{table}

\begin{table}[t]
\small
\centering
\begin{tabular}{c|c|c|c}
\hline
    Viewpoint Pool &  Aligned  &  {Hemispherical}  &  {Spherical} \\
\hline
    with                    &\textbf{65.52\%}  &\textbf{53.53\%}  &\textbf{46.12\%}    \\
    without                 &65.28\%  &52.07\%  &42.39\% \\
\hline
\end{tabular}
\caption{The ablation study on whether the viewpoint pool is used in the proposed method. With the viewpoint pool, the method achieves much better mIoU results.}
\label{table:ablation_viewPool}
\end{table}

\begin{figure*}[t]
\begin{center}
   \includegraphics[width=\linewidth]{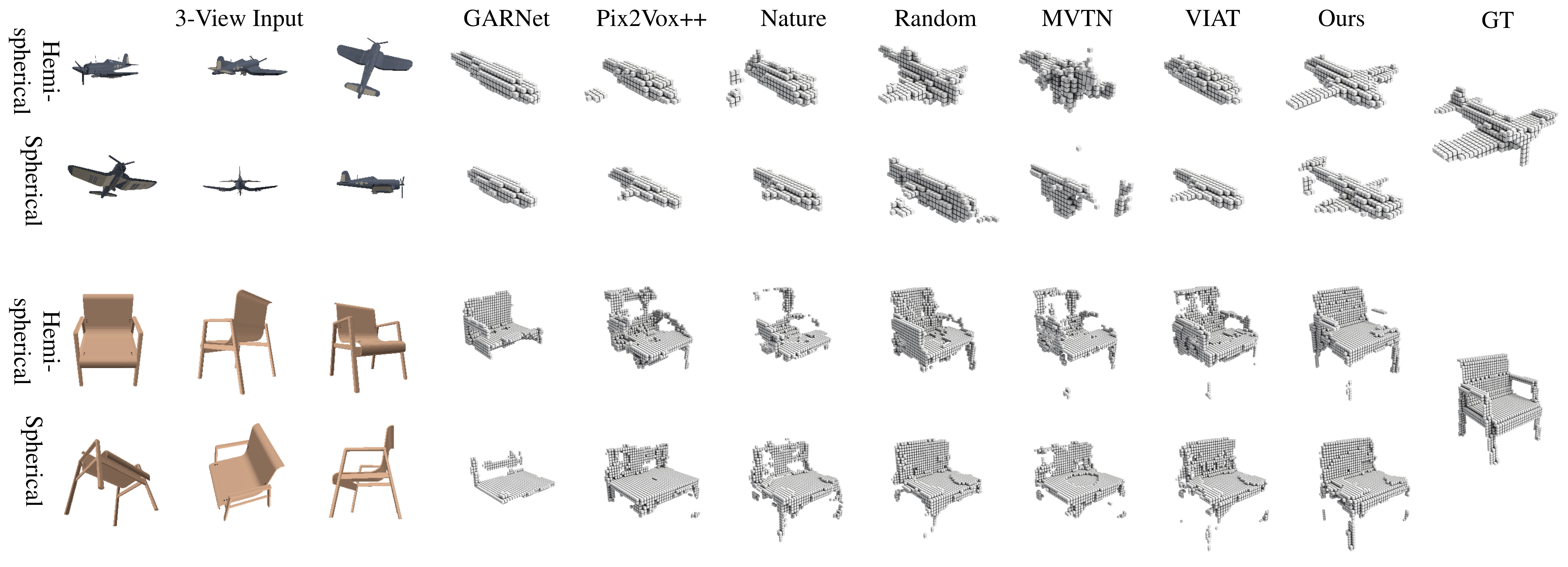}
\end{center}
\caption{The visualization results of CNNs-based methods, and we use Pix2Vox++ as the 3D reconstruction model.}
\label{fig:Pix2Vox++}
\end{figure*}

\begin{figure*}[t]
\begin{center}
   \includegraphics[width=\linewidth]{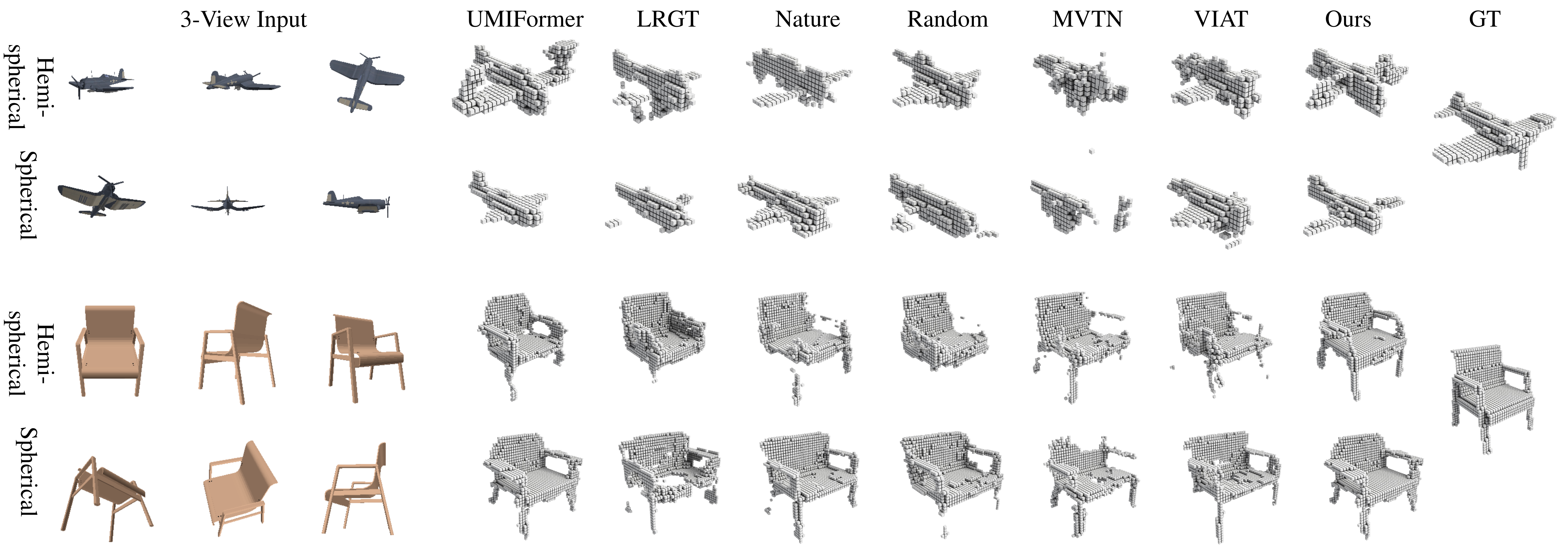}
\end{center}
\caption{The visualization results of Transformer-based methods, and we use LRGT as the 3D reconstruction model.}
\label{fig:LRGT}
\end{figure*}

Table \ref{table:MV_results_transformer} shows the VTR of Transformer-based multi-view 3D object reconstruction methods. VTR methods utilize LRGT \cite{yang2023long} as the 3D reconstruction model. Similarly, the proposed method also outperforms all Transformer SOTAs  \cite{zhu2023umiformer,yang2023long} (None-VTR) and view transformation robustness comparisons (VTR) on `Hemispherical' and `Spherical', which proves the advantage of the proposed method and its effectiveness of boosting the robustness of Transfomer-based 3D reconstruction methods.
The performance gain of our method over comparisons is larger for CNNs-based 3D reconstruction models, due to the weak view fusion ability of CNNs models compared to Transformer 3D reconstruction models.
\textbf{Overall, from Table \ref{table:MV_results_CNNs} and \ref{table:MV_results_transformer}, we can conclude the proposed method could boost the robustness of both CNNs-based and Transformer-based 3D reconstruction models, indicating our method's generalization ability}. 

\textbf{Visualization results.} 
We show the visualization examples of the proposed method and comparisons in Figure \ref{fig:Pix2Vox++} (CNNs-based) and Figure \ref{fig:LRGT} (Transformer-based).
In each figure, for each object example, the left three columns show the input 3 views, and the middle seven columns indicate the voxel predictions of the methods, and the last column is the ground-truth voxel. The first and the second row of each object example represent results on `Hemispherical' and `Spherical', respectively.
In Figure \ref{fig:Pix2Vox++} and Figure \ref{fig:LRGT}, our method can recover the rough shape of the plane and the chair, while comparisons ignore a lot of details, like the wings of the plane in Figure \ref{fig:Pix2Vox++} and the chair legs in Figure \ref{fig:LRGT}. 
We can also observe that the results of the same object of each method on `Spherical' are worse than on `Hemispherical', because `Spherical' contains larger view transformations and is more difficult. Transformer-based methods can achieve better visualization results than CNN-based methods because Transformer methods have stronger inter-view fusion ability. 
No matter whether using which 3D reconstruction model or on which testing set, the proposed method always achieves the best results, which indicates the stronger robustness of the proposed method to view transformations.

\begin{table}[t]
\scriptsize
\centering
\begin{tabular}
{@{\hspace{0.02cm}}l@{\hspace{0.12cm}}|c@{\hspace{0.12cm}}c@{\hspace{0.12cm}}|c@{\hspace{0.12cm}}c@{\hspace{0.12cm}}|c@{\hspace{0.12cm}}c@{\hspace{0.02cm}}}
\hline
    Test set &  \multicolumn{2}{c|}{Aligned}  &  \multicolumn{2}{c|}{Hemispherical}  &  \multicolumn{2}{c}{Spherical} \\
    Method                                     & mIoU    & F\_score       & mIoU    & F\_score        & mIoU    & F\_score  \\
\hline
    OCCNet                   & 52.89\%    & 33.46\%        & 38.31 \%   &  21.49\%  & 26.62\%  & 13.74\%    \\ 
    Pix2Vox++                    & 56.30\%    & 38.93\%        & 39.12\%   & 23.90\%  & 30.06\%  & 17.39\%    \\ 
    
    LRGT                    & 62.62\%    & 42.63\%        & 45.17\%   &   26.63\%  & 34.70\%  & 19.52\%    \\ 
\hline
    Ours (Pix2Vox++)                   & 59.47\%    & 40.32\%        &  43.81\%     &  25.84\%  & \textbf{36.54}\%       & \textbf{20.71}\%        \\
    Ours (LRGT)                         & \textbf{63.21}\%    & \textbf{43.35}\%        &  \textbf{47.14}\%     &  \textbf{28.12}\%  & 36.13\%       & 20.20\%        \\
\hline
\end{tabular}
\caption{The experiments on single-view 3D object reconstruction methods on the test set with view transformations.
}
\label{table:SingleView_results}
\end{table}

\begin{table}[t]
\small
\centering
\begin{tabular}{l|cc}
\hline
    Method &  mIoU  &  F\_score   \\
\hline
    Pix2Vox++           & 16.24\%    & 7.76\%    \\
    Ours (Pix2Vox++)     & \textbf{16.45\%}    & \textbf{7.77\%}    \\
\hline
    LRGT                & 17.08\%    & 7.79\%    \\
    Ours (LRGT)          & \textbf{17.47\%}    & \textbf{8.59\%}    \\
\hline
\end{tabular}
\caption{The testing results on the Pix3D Chairs dataset.}
\label{table:chair_results}
\end{table}

\subsubsection{Ablation study} 
\textbf{Ablation study on the interval of degree $K$.}
We conduct the ablation study on the interval of degree $K$ in the view selection module. 
Table \ref{table:ablation_angle} presents the reconstruction result of the model after iterative training with different $K$, where the best results are achieved by $K=30^{\circ}$. It shows a suitable interval of degree benefits the model in achieving better results. If the degree is too large, the viewpoint selection will tend to be random; while if it is too small, it will affect the viewpoint diversity and hinder the reconstruction robustness. Thus, we experiment with $K=30^{\circ}$ in our methods.

\textbf{Ablation study on the viewpoint pool.} 
We conduct the ablation study on the effectiveness of the viewpoint pool in Table \ref{table:ablation_viewPool}. 
It shows that the method with the viewpoint pool can achieve better performance than without the viewpoint pool on all testing sets, especially on the `Spherical' set with much larger view transformations. This indicates the effectiveness of the viewpoint pool in boosting the multi-view 3D reconstruction model's robustness to view transformations.

\textbf{Single-view 3D reconstruction performance.}
We also validate the proposed method's robustness in single-view 3D reconstruction and compare it with single-view SOTA methods \cite{mescheder2019occupancy,xie2020pix2vox++,yang2023long} in Table \ref{table:SingleView_results}.
It shows that our method achieves better results than all comparisons on all sets, no matter whether using Pix2Vox++ \cite{xie2020pix2vox++}   or  LRGT \cite{yang2023long} as 3D reconstruction models,  which proves the effectiveness of the proposed method for single-view 3D reconstruction.

\textbf{Evaluation on the real-world Dataset Pix3D.}
We also test the proposed method on Pix3D \cite{sun2018pix3d} to verify its single-view reconstruction performance on real-world data with more complicated object viewpoint distributions.
Following \cite{xie2020pix2vox++}, we use the data from the Chair category in ShapeNet to generate a training set and render images with random backgrounds in the SUN dataset \cite{xiao2010sun}, and each object has 60 synthesized images. Table \ref{table:chair_results} shows that our method's performance is better than SOTA methods, suggesting that our method helps the 3D reconstruction model generalize better to real-world datasets by improving the model's view transformation robustness.

\section{Conclusion}
In this paper, we propose a novel reconstruction error-guided view selection method together with view synthesis via a Stable Diffusion model to improve the view transformation robustness of existing multi-view 3D object reconstruction methods. 
Instead of randomly synthesizing new views from Stable Diffusion models and adding them to model training as a data augmentation approach, we consider the spatial distributions of the 3D reconstruction errors and use them to guide the view selection process for choosing the most effective views covering the errors as most as possible. Our proposed method shows the best view transformation robustness compared to the latest multi-view 3D reconstruction SOTAs and various view transformation robustness comparison methods. 
We provide a new perspective on incorporating large vision models into existing relatively `small' 3D object reconstruction models for robustness gains without increasing the model deploying cost.

\section{Acknowledgements}
This work was supported in parts by NSFC (62202312, U21B2023), Guangdong Basic and Applied Basic Research Foundation (2023B1515120026), Shenzhen Science and Technology Program (KQTD 20210811090044003, RCJC20200714114435012), and Scientific Development Funds from Shenzhen University.

\bibliography{aaai25}


\begin{thebibliography}{46}
\ifx \bisbn   \undefined \def \bisbn  #1{ISBN #1}\fi
\ifx \binits  \undefined \def \binits#1{#1}\fi
\ifx \bauthor  \undefined \def \bauthor#1{#1}\fi
\ifx \batitle  \undefined \def \batitle#1{#1}\fi
\ifx \bjtitle  \undefined \def \bjtitle#1{#1}\fi
\ifx \bvolume  \undefined \def \bvolume#1{\textbf{#1}}\fi
\ifx \byear  \undefined \def \byear#1{#1}\fi
\ifx \bissue  \undefined \def \bissue#1{#1}\fi
\ifx \bfpage  \undefined \def \bfpage#1{#1}\fi
\ifx \blpage  \undefined \def \blpage #1{#1}\fi
\ifx \burl  \undefined \def \burl#1{\textsf{#1}}\fi
\ifx \doiurl  \undefined \def \doiurl#1{\url{https://doi.org/#1}}\fi
\ifx \betal  \undefined \def \betal{\textit{et al.}}\fi
\ifx \binstitute  \undefined \def \binstitute#1{#1}\fi
\ifx \binstitutionaled  \undefined \def \binstitutionaled#1{#1}\fi
\ifx \bctitle  \undefined \def \bctitle#1{#1}\fi
\ifx \beditor  \undefined \def \beditor#1{#1}\fi
\ifx \bpublisher  \undefined \def \bpublisher#1{#1}\fi
\ifx \bbtitle  \undefined \def \bbtitle#1{#1}\fi
\ifx \bedition  \undefined \def \bedition#1{#1}\fi
\ifx \bseriesno  \undefined \def \bseriesno#1{#1}\fi
\ifx \blocation  \undefined \def \blocation#1{#1}\fi
\ifx \bsertitle  \undefined \def \bsertitle#1{#1}\fi
\ifx \bsnm \undefined \def \bsnm#1{#1}\fi
\ifx \bsuffix \undefined \def \bsuffix#1{#1}\fi
\ifx \bparticle \undefined \def \bparticle#1{#1}\fi
\ifx \barticle \undefined \def \barticle#1{#1}\fi
\bibcommenthead
\ifx \bconfdate \undefined \def \bconfdate #1{#1}\fi
\ifx \botherref \undefined \def \botherref #1{#1}\fi
\ifx \url \undefined \def \url#1{\textsf{#1}}\fi
\ifx \bchapter \undefined \def \bchapter#1{#1}\fi
\ifx \bbook \undefined \def \bbook#1{#1}\fi
\ifx \bcomment \undefined \def \bcomment#1{#1}\fi
\ifx \oauthor \undefined \def \oauthor#1{#1}\fi
\ifx \citeauthoryear \undefined \def \citeauthoryear#1{#1}\fi
\ifx \endbibitem  \undefined \def \endbibitem {}\fi
\ifx \bconflocation  \undefined \def \bconflocation#1{#1}\fi
\ifx \arxivurl  \undefined \def \arxivurl#1{\textsf{#1}}\fi
\csname PreBibitemsHook\endcsname

\bibitem[\protect\citeauthoryear{Wei et~al.}{2020}]{wei2020view}
\begin{bchapter}
\bauthor{\bsnm{Wei}, \binits{X.}},
\bauthor{\bsnm{Yu}, \binits{R.}},
\bauthor{\bsnm{Sun}, \binits{J.}}:
\bctitle{View-gcn: View-based graph convolutional network for 3d shape analysis}.
In: \bbtitle{Proceedings of the IEEE/CVF Conference on Computer Vision and Pattern Recognition},
pp. \bfpage{1850}--\blpage{1859}
(\byear{2020})
\end{bchapter}
\endbibitem

\bibitem[\protect\citeauthoryear{Qi et~al.}{2017}]{qi2017pointnet}
\begin{bchapter}
\bauthor{\bsnm{Qi}, \binits{C.R.}},
\bauthor{\bsnm{Su}, \binits{H.}},
\bauthor{\bsnm{Mo}, \binits{K.}},
\bauthor{\bsnm{Guibas}, \binits{L.J.}}:
\bctitle{Pointnet: Deep learning on point sets for 3d classification and segmentation}.
In: \bbtitle{Proceedings of the IEEE Conference on Computer Vision and Pattern Recognition},
pp. \bfpage{652}--\blpage{660}
(\byear{2017})
\end{bchapter}
\endbibitem

\bibitem[\protect\citeauthoryear{Wang et~al.}{2022}]{Wang2022ARTPointIR}
\begin{botherref}
\oauthor{\bsnm{Wang}, \binits{R.}},
\oauthor{\bsnm{Yang}, \binits{Y.}},
\oauthor{\bsnm{Tao}, \binits{D.}}:
Art-point: Improving rotation robustness of point cloud classifiers via adversarial rotation.
2022 IEEE/CVF Conference on Computer Vision and Pattern Recognition (CVPR),
14351--14360
(2022)
\end{botherref}
\endbibitem

\bibitem[\protect\citeauthoryear{Rombach et~al.}{2022}]{rombach2022high}
\begin{bchapter}
\bauthor{\bsnm{Rombach}, \binits{R.}},
\bauthor{\bsnm{Blattmann}, \binits{A.}},
\bauthor{\bsnm{Lorenz}, \binits{D.}},
\bauthor{\bsnm{Esser}, \binits{P.}},
\bauthor{\bsnm{Ommer}, \binits{B.}}:
\bctitle{High-resolution image synthesis with latent diffusion models}.
In: \bbtitle{Proceedings of the IEEE/CVF Conference on Computer Vision and Pattern Recognition},
pp. \bfpage{10684}--\blpage{10695}
(\byear{2022})
\end{bchapter}
\endbibitem

\bibitem[\protect\citeauthoryear{Hu et~al.}{2024}]{hu2024neural}
\begin{barticle}
\bauthor{\bsnm{Hu}, \binits{J.}},
\bauthor{\bsnm{Hui}, \binits{K.-H.}},
\bauthor{\bsnm{Liu}, \binits{Z.}},
\bauthor{\bsnm{Li}, \binits{R.}},
\bauthor{\bsnm{Fu}, \binits{C.-W.}}:
\batitle{Neural wavelet-domain diffusion for 3d shape generation, inversion, and manipulation}.
\bjtitle{ACM Transactions on Graphics}
\bvolume{43}(\bissue{2}),
\bfpage{1}--\blpage{18}
(\byear{2024})
\end{barticle}
\endbibitem

\bibitem[\protect\citeauthoryear{Shi et~al.}{2023}]{shi2023zero123++}
\begin{botherref}
\oauthor{\bsnm{Shi}, \binits{R.}},
\oauthor{\bsnm{Chen}, \binits{H.}},
\oauthor{\bsnm{Zhang}, \binits{Z.}},
\oauthor{\bsnm{Liu}, \binits{M.}},
\oauthor{\bsnm{Xu}, \binits{C.}},
\oauthor{\bsnm{Wei}, \binits{X.}},
\oauthor{\bsnm{Chen}, \binits{L.}},
\oauthor{\bsnm{Zeng}, \binits{C.}},
\oauthor{\bsnm{Su}, \binits{H.}}:
Zero123++: a single image to consistent multi-view diffusion base model.
arXiv preprint arXiv:2310.15110
(2023)
\end{botherref}
\endbibitem

\bibitem[\protect\citeauthoryear{Qi et~al.}{2024}]{qi2024vpp}
\begin{botherref}
\oauthor{\bsnm{Qi}, \binits{Z.}},
\oauthor{\bsnm{Yu}, \binits{M.}},
\oauthor{\bsnm{Dong}, \binits{R.}},
\oauthor{\bsnm{Ma}, \binits{K.}}:
Vpp: Efficient conditional 3d generation via voxel-point progressive representation.
Advances in Neural Information Processing Systems
\textbf{36}
(2024)
\end{botherref}
\endbibitem

\bibitem[\protect\citeauthoryear{Yoo et~al.}{2024}]{yoo2024dreamsparse}
\begin{botherref}
\oauthor{\bsnm{Yoo}, \binits{P.}},
\oauthor{\bsnm{Guo}, \binits{J.}},
\oauthor{\bsnm{Matsuo}, \binits{Y.}},
\oauthor{\bsnm{Gu}, \binits{S.S.}}:
Dreamsparse: Escaping from plato’s cave with 2d diffusion model given sparse views.
Advances in Neural Information Processing Systems
\textbf{36}
(2024)
\end{botherref}
\endbibitem

\bibitem[\protect\citeauthoryear{Zou et~al.}{2024}]{zou2024sparse3d}
\begin{bchapter}
\bauthor{\bsnm{Zou}, \binits{Z.}},
\bauthor{\bsnm{Cheng}, \binits{W.}},
\bauthor{\bsnm{Cao}, \binits{Y.-P.}},
\bauthor{\bsnm{Huang}, \binits{S.-S.}},
\bauthor{\bsnm{Shan}, \binits{Y.}},
\bauthor{\bsnm{Zhang}, \binits{S.-H.}}:
\bctitle{Sparse3d: Distilling multiview-consistent diffusion for object reconstruction from sparse views}.
In: \bbtitle{Proceedings of the AAAI Conference on Artificial Intelligence},
vol. \bseriesno{38},
pp. \bfpage{7900}--\blpage{7908}
(\byear{2024})
\end{bchapter}
\endbibitem

\bibitem[\protect\citeauthoryear{Kwak et~al.}{2024}]{kwak2024vivid}
\begin{bchapter}
\bauthor{\bsnm{Kwak}, \binits{J.-g.}},
\bauthor{\bsnm{Dong}, \binits{E.}},
\bauthor{\bsnm{Jin}, \binits{Y.}},
\bauthor{\bsnm{Ko}, \binits{H.}},
\bauthor{\bsnm{Mahajan}, \binits{S.}},
\bauthor{\bsnm{Yi}, \binits{K.M.}}:
\bctitle{Vivid-1-to-3: Novel view synthesis with video diffusion models}.
In: \bbtitle{Proceedings of the IEEE/CVF Conference on Computer Vision and Pattern Recognition},
pp. \bfpage{6775}--\blpage{6785}
(\byear{2024})
\end{bchapter}
\endbibitem

\bibitem[\protect\citeauthoryear{Bauer et~al.}{2024}]{bauer2024challenges}
\begin{botherref}
\oauthor{\bsnm{Bauer}, \binits{D.}},
\oauthor{\bsnm{H{\"o}nig}, \binits{P.}},
\oauthor{\bsnm{Weibel}, \binits{J.-B.}},
\oauthor{\bsnm{Garc{\'\i}a-Rodr{\'\i}guez}, \binits{J.}},
\oauthor{\bsnm{Vincze}, \binits{M.}}, et al.:
Challenges for monocular 6d object pose estimation in robotics.
IEEE Transactions on Robotics
(2024)
\end{botherref}
\endbibitem

\bibitem[\protect\citeauthoryear{Liu et~al.}{2024}]{liu2024one}
\begin{botherref}
\oauthor{\bsnm{Liu}, \binits{M.}},
\oauthor{\bsnm{Xu}, \binits{C.}},
\oauthor{\bsnm{Jin}, \binits{H.}},
\oauthor{\bsnm{Chen}, \binits{L.}},
\oauthor{\bsnm{Varma~T}, \binits{M.}},
\oauthor{\bsnm{Xu}, \binits{Z.}},
\oauthor{\bsnm{Su}, \binits{H.}}:
One-2-3-45: Any single image to 3d mesh in 45 seconds without per-shape optimization.
Advances in Neural Information Processing Systems
\textbf{36}
(2024)
\end{botherref}
\endbibitem

\bibitem[\protect\citeauthoryear{Liu et~al.}{2023a}]{liu2023syncdreamer}
\begin{botherref}
\oauthor{\bsnm{Liu}, \binits{Y.}},
\oauthor{\bsnm{Lin}, \binits{C.}},
\oauthor{\bsnm{Zeng}, \binits{Z.}},
\oauthor{\bsnm{Long}, \binits{X.}},
\oauthor{\bsnm{Liu}, \binits{L.}},
\oauthor{\bsnm{Komura}, \binits{T.}},
\oauthor{\bsnm{Wang}, \binits{W.}}:
Syncdreamer: Generating multiview-consistent images from a single-view image.
arXiv preprint arXiv:2309.03453
(2023)
\end{botherref}
\endbibitem

\bibitem[\protect\citeauthoryear{Liu et~al.}{2023b}]{liu2023one2345++}
\begin{botherref}
\oauthor{\bsnm{Liu}, \binits{M.}},
\oauthor{\bsnm{Shi}, \binits{R.}},
\oauthor{\bsnm{Chen}, \binits{L.}},
\oauthor{\bsnm{Zhang}, \binits{Z.}},
\oauthor{\bsnm{Xu}, \binits{C.}},
\oauthor{\bsnm{Wei}, \binits{X.}},
\oauthor{\bsnm{Chen}, \binits{H.}},
\oauthor{\bsnm{Zeng}, \binits{C.}},
\oauthor{\bsnm{Gu}, \binits{J.}},
\oauthor{\bsnm{Su}, \binits{H.}}:
One-2-3-45++: Fast single image to 3d objects with consistent multi-view generation and 3d diffusion.
arXiv preprint arXiv:2311.07885
(2023)
\end{botherref}
\endbibitem

\bibitem[\protect\citeauthoryear{Liu et~al.}{2023c}]{Liu2023Zero1to3ZO}
\begin{botherref}
\oauthor{\bsnm{Liu}, \binits{R.}},
\oauthor{\bsnm{Wu}, \binits{R.}},
\oauthor{\bsnm{Hoorick}, \binits{B.V.}},
\oauthor{\bsnm{Tokmakov}, \binits{P.}},
\oauthor{\bsnm{Zakharov}, \binits{S.}},
\oauthor{\bsnm{Vondrick}, \binits{C.}}:
Zero-1-to-3: Zero-shot one image to 3d object.
2023 IEEE/CVF International Conference on Computer Vision (ICCV),
9264--9275
(2023)
\end{botherref}
\endbibitem

\bibitem[\protect\citeauthoryear{Xie et~al.}{2020}]{xie2020pix2vox++}
\begin{barticle}
\bauthor{\bsnm{Xie}, \binits{H.}},
\bauthor{\bsnm{Yao}, \binits{H.}},
\bauthor{\bsnm{Zhang}, \binits{S.}},
\bauthor{\bsnm{Zhou}, \binits{S.}},
\bauthor{\bsnm{Sun}, \binits{W.}}:
\batitle{Pix2vox++: Multi-scale context-aware 3d object reconstruction from single and multiple images}.
\bjtitle{International Journal of Computer Vision}
\bvolume{128}(\bissue{12}),
\bfpage{2919}--\blpage{2935}
(\byear{2020})
\end{barticle}
\endbibitem

\bibitem[\protect\citeauthoryear{Yang et~al.}{2023}]{yang2023long}
\begin{bchapter}
\bauthor{\bsnm{Yang}, \binits{L.}},
\bauthor{\bsnm{Zhu}, \binits{Z.}},
\bauthor{\bsnm{Lin}, \binits{X.}},
\bauthor{\bsnm{Nong}, \binits{J.}},
\bauthor{\bsnm{Liang}, \binits{Y.}}:
\bctitle{Long-range grouping transformer for multi-view 3d reconstruction}.
In: \bbtitle{Proceedings of the IEEE/CVF International Conference on Computer Vision},
pp. \bfpage{18257}--\blpage{18267}
(\byear{2023})
\end{bchapter}
\endbibitem

\bibitem[\protect\citeauthoryear{Chang et~al.}{2015}]{chang2015shapenet}
\begin{botherref}
\oauthor{\bsnm{Chang}, \binits{A.X.}},
\oauthor{\bsnm{Funkhouser}, \binits{T.}},
\oauthor{\bsnm{Guibas}, \binits{L.}},
\oauthor{\bsnm{Hanrahan}, \binits{P.}},
\oauthor{\bsnm{Huang}, \binits{Q.}},
\oauthor{\bsnm{Li}, \binits{Z.}},
\oauthor{\bsnm{Savarese}, \binits{S.}},
\oauthor{\bsnm{Savva}, \binits{M.}},
\oauthor{\bsnm{Song}, \binits{S.}},
\oauthor{\bsnm{Su}, \binits{H.}}, et al.:
Shapenet: An information-rich 3d model repository.
arXiv preprint arXiv:1512.03012
(2015)
\end{botherref}
\endbibitem

\bibitem[\protect\citeauthoryear{Su et~al.}{2015}]{su2015multi}
\begin{bchapter}
\bauthor{\bsnm{Su}, \binits{H.}},
\bauthor{\bsnm{Maji}, \binits{S.}},
\bauthor{\bsnm{Kalogerakis}, \binits{E.}},
\bauthor{\bsnm{Learned-Miller}, \binits{E.}}:
\bctitle{Multi-view convolutional neural networks for 3d shape recognition}.
In: \bbtitle{Proceedings of the IEEE International Conference on Computer Vision},
pp. \bfpage{945}--\blpage{953}
(\byear{2015})
\end{bchapter}
\endbibitem

\bibitem[\protect\citeauthoryear{Paschalidou et~al.}{2018}]{paschalidou2018raynet}
\begin{bchapter}
\bauthor{\bsnm{Paschalidou}, \binits{D.}},
\bauthor{\bsnm{Ulusoy}, \binits{O.}},
\bauthor{\bsnm{Schmitt}, \binits{C.}},
\bauthor{\bsnm{Van~Gool}, \binits{L.}},
\bauthor{\bsnm{Geiger}, \binits{A.}}:
\bctitle{Raynet: Learning volumetric 3d reconstruction with ray potentials}.
In: \bbtitle{Proceedings of the IEEE Conference on Computer Vision and Pattern Recognition},
pp. \bfpage{3897}--\blpage{3906}
(\byear{2018})
\end{bchapter}
\endbibitem

\bibitem[\protect\citeauthoryear{Huang et~al.}{2018}]{huang2018deepmvs}
\begin{bchapter}
\bauthor{\bsnm{Huang}, \binits{P.-H.}},
\bauthor{\bsnm{Matzen}, \binits{K.}},
\bauthor{\bsnm{Kopf}, \binits{J.}},
\bauthor{\bsnm{Ahuja}, \binits{N.}},
\bauthor{\bsnm{Huang}, \binits{J.-B.}}:
\bctitle{Deepmvs: Learning multi-view stereopsis}.
In: \bbtitle{Proceedings of the IEEE Conference on Computer Vision and Pattern Recognition},
pp. \bfpage{2821}--\blpage{2830}
(\byear{2018})
\end{bchapter}
\endbibitem

\bibitem[\protect\citeauthoryear{Yao et~al.}{2018}]{yao2018mvsnet}
\begin{bchapter}
\bauthor{\bsnm{Yao}, \binits{Y.}},
\bauthor{\bsnm{Luo}, \binits{Z.}},
\bauthor{\bsnm{Li}, \binits{S.}},
\bauthor{\bsnm{Fang}, \binits{T.}},
\bauthor{\bsnm{Quan}, \binits{L.}}:
\bctitle{Mvsnet: Depth inference for unstructured multi-view stereo}.
In: \bbtitle{Proceedings of the European Conference on Computer Vision (ECCV)},
pp. \bfpage{767}--\blpage{783}
(\byear{2018})
\end{bchapter}
\endbibitem

\bibitem[\protect\citeauthoryear{Choy et~al.}{2016}]{choy20163d}
\begin{bchapter}
\bauthor{\bsnm{Choy}, \binits{C.B.}},
\bauthor{\bsnm{Xu}, \binits{D.}},
\bauthor{\bsnm{Gwak}, \binits{J.}},
\bauthor{\bsnm{Chen}, \binits{K.}},
\bauthor{\bsnm{Savarese}, \binits{S.}}:
\bctitle{3d-r2n2: A unified approach for single and multi-view 3d object reconstruction}.
In: \bbtitle{Computer Vision--ECCV 2016: 14th European Conference, Amsterdam, The Netherlands, October 11-14, 2016, Proceedings, Part VIII 14},
pp. \bfpage{628}--\blpage{644}
(\byear{2016}).
\bcomment{Springer}
\end{bchapter}
\endbibitem

\bibitem[\protect\citeauthoryear{Kar et~al.}{2017}]{kar2017learning}
\begin{botherref}
\oauthor{\bsnm{Kar}, \binits{A.}},
\oauthor{\bsnm{H{\"a}ne}, \binits{C.}},
\oauthor{\bsnm{Malik}, \binits{J.}}:
Learning a multi-view stereo machine.
Advances in neural information processing systems
\textbf{30}
(2017)
\end{botherref}
\endbibitem

\bibitem[\protect\citeauthoryear{Dosovitskiy et~al.}{2020}]{dosovitskiy2020image}
\begin{botherref}
\oauthor{\bsnm{Dosovitskiy}, \binits{A.}},
\oauthor{\bsnm{Beyer}, \binits{L.}},
\oauthor{\bsnm{Kolesnikov}, \binits{A.}},
\oauthor{\bsnm{Weissenborn}, \binits{D.}},
\oauthor{\bsnm{Zhai}, \binits{X.}},
\oauthor{\bsnm{Unterthiner}, \binits{T.}},
\oauthor{\bsnm{Dehghani}, \binits{M.}},
\oauthor{\bsnm{Minderer}, \binits{M.}},
\oauthor{\bsnm{Heigold}, \binits{G.}},
\oauthor{\bsnm{Gelly}, \binits{S.}}, et al.:
An image is worth 16x16 words: Transformers for image recognition at scale.
arXiv preprint arXiv:2010.11929
(2020)
\end{botherref}
\endbibitem

\bibitem[\protect\citeauthoryear{Yao et~al.}{2019}]{yao2019recurrent}
\begin{bchapter}
\bauthor{\bsnm{Yao}, \binits{Y.}},
\bauthor{\bsnm{Luo}, \binits{Z.}},
\bauthor{\bsnm{Li}, \binits{S.}},
\bauthor{\bsnm{Shen}, \binits{T.}},
\bauthor{\bsnm{Fang}, \binits{T.}},
\bauthor{\bsnm{Quan}, \binits{L.}}:
\bctitle{Recurrent mvsnet for high-resolution multi-view stereo depth inference}.
In: \bbtitle{Proceedings of the IEEE/CVF Conference on Computer Vision and Pattern Recognition},
pp. \bfpage{5525}--\blpage{5534}
(\byear{2019})
\end{bchapter}
\endbibitem

\bibitem[\protect\citeauthoryear{Yang et~al.}{2020}]{yang2020robust}
\begin{barticle}
\bauthor{\bsnm{Yang}, \binits{B.}},
\bauthor{\bsnm{Wang}, \binits{S.}},
\bauthor{\bsnm{Markham}, \binits{A.}},
\bauthor{\bsnm{Trigoni}, \binits{N.}}:
\batitle{Robust attentional aggregation of deep feature sets for multi-view 3d reconstruction}.
\bjtitle{International Journal of Computer Vision}
\bvolume{128}(\bissue{1}),
\bfpage{53}--\blpage{73}
(\byear{2020})
\end{barticle}
\endbibitem

\bibitem[\protect\citeauthoryear{Xie et~al.}{2019}]{xie2019pix2vox}
\begin{bchapter}
\bauthor{\bsnm{Xie}, \binits{H.}},
\bauthor{\bsnm{Yao}, \binits{H.}},
\bauthor{\bsnm{Sun}, \binits{X.}},
\bauthor{\bsnm{Zhou}, \binits{S.}},
\bauthor{\bsnm{Zhang}, \binits{S.}}:
\bctitle{Pix2vox: Context-aware 3d reconstruction from single and multi-view images}.
In: \bbtitle{Proceedings of the IEEE/CVF International Conference on Computer Vision},
pp. \bfpage{2690}--\blpage{2698}
(\byear{2019})
\end{bchapter}
\endbibitem

\bibitem[\protect\citeauthoryear{Zhu et~al.}{2023}]{zhu2023garnet}
\begin{barticle}
\bauthor{\bsnm{Zhu}, \binits{Z.}},
\bauthor{\bsnm{Yang}, \binits{L.}},
\bauthor{\bsnm{Lin}, \binits{X.}},
\bauthor{\bsnm{Yang}, \binits{L.}},
\bauthor{\bsnm{Liang}, \binits{Y.}}:
\batitle{Garnet: Global-aware multi-view 3d reconstruction network and the cost-performance tradeoff}.
\bjtitle{Pattern Recognition}
\bvolume{142},
\bfpage{109674}
(\byear{2023})
\end{barticle}
\endbibitem

\bibitem[\protect\citeauthoryear{Shi et~al.}{2021}]{shi20213d}
\begin{botherref}
\oauthor{\bsnm{Shi}, \binits{Z.}},
\oauthor{\bsnm{Meng}, \binits{Z.}},
\oauthor{\bsnm{Xing}, \binits{Y.}},
\oauthor{\bsnm{Ma}, \binits{Y.}},
\oauthor{\bsnm{Wattenhofer}, \binits{R.}}:
3d-retr: End-to-end single and multi-view 3d reconstruction with transformers.
arXiv preprint arXiv:2110.08861
(2021)
\end{botherref}
\endbibitem

\bibitem[\protect\citeauthoryear{Tiong et~al.}{2022}]{tiong20223d}
\begin{bchapter}
\bauthor{\bsnm{Tiong}, \binits{L.C.O.}},
\bauthor{\bsnm{Sigmund}, \binits{D.}},
\bauthor{\bsnm{Teoh}, \binits{A.B.J.}}:
\bctitle{3d-c2ft: Coarse-to-fine transformer for multi-view 3d reconstruction}.
In: \bbtitle{Proceedings of the Asian Conference on Computer Vision},
pp. \bfpage{1438}--\blpage{1454}
(\byear{2022})
\end{bchapter}
\endbibitem

\bibitem[\protect\citeauthoryear{Wang et~al.}{2021}]{wang2021multi}
\begin{bchapter}
\bauthor{\bsnm{Wang}, \binits{D.}},
\bauthor{\bsnm{Cui}, \binits{X.}},
\bauthor{\bsnm{Chen}, \binits{X.}},
\bauthor{\bsnm{Zou}, \binits{Z.}},
\bauthor{\bsnm{Shi}, \binits{T.}},
\bauthor{\bsnm{Salcudean}, \binits{S.}},
\bauthor{\bsnm{Wang}, \binits{Z.J.}},
\bauthor{\bsnm{Ward}, \binits{R.}}:
\bctitle{Multi-view 3d reconstruction with transformers}.
In: \bbtitle{Proceedings of the IEEE/CVF International Conference on Computer Vision},
pp. \bfpage{5722}--\blpage{5731}
(\byear{2021})
\end{bchapter}
\endbibitem

\bibitem[\protect\citeauthoryear{Yagubbayli et~al.}{2021}]{yagubbayli2021legoformer}
\begin{botherref}
\oauthor{\bsnm{Yagubbayli}, \binits{F.}},
\oauthor{\bsnm{Wang}, \binits{Y.}},
\oauthor{\bsnm{Tonioni}, \binits{A.}},
\oauthor{\bsnm{Tombari}, \binits{F.}}:
Legoformer: Transformers for block-by-block multi-view 3d reconstruction.
arXiv preprint arXiv:2106.12102
(2021)
\end{botherref}
\endbibitem

\bibitem[\protect\citeauthoryear{Arshad and Beksi}{2023}]{arshad2023list}
\begin{bchapter}
\bauthor{\bsnm{Arshad}, \binits{M.S.}},
\bauthor{\bsnm{Beksi}, \binits{W.J.}}:
\bctitle{List: Learning implicitly from spatial transformers for single-view 3d reconstruction}.
In: \bbtitle{Proceedings of the IEEE/CVF International Conference on Computer Vision},
pp. \bfpage{9321}--\blpage{9330}
(\byear{2023})
\end{bchapter}
\endbibitem

\bibitem[\protect\citeauthoryear{Zhu et~al.}{2023}]{zhu2023umiformer}
\begin{bchapter}
\bauthor{\bsnm{Zhu}, \binits{Z.}},
\bauthor{\bsnm{Yang}, \binits{L.}},
\bauthor{\bsnm{Li}, \binits{N.}},
\bauthor{\bsnm{Jiang}, \binits{C.}},
\bauthor{\bsnm{Liang}, \binits{Y.}}:
\bctitle{Umiformer: Mining the correlations between similar tokens for multi-view 3d reconstruction}.
In: \bbtitle{Proceedings of the IEEE/CVF International Conference on Computer Vision},
pp. \bfpage{18226}--\blpage{18235}
(\byear{2023})
\end{bchapter}
\endbibitem

\bibitem[\protect\citeauthoryear{Spezialetti et~al.}{2019}]{spezialetti2019learning}
\begin{bchapter}
\bauthor{\bsnm{Spezialetti}, \binits{R.}},
\bauthor{\bsnm{Salti}, \binits{S.}},
\bauthor{\bsnm{Stefano}, \binits{L.D.}}:
\bctitle{Learning an effective equivariant 3d descriptor without supervision}.
In: \bbtitle{Proceedings of the IEEE/CVF International Conference on Computer Vision},
pp. \bfpage{6401}--\blpage{6410}
(\byear{2019})
\end{bchapter}
\endbibitem

\bibitem[\protect\citeauthoryear{Zhu et~al.}{2023}]{zhu2023r}
\begin{botherref}
\oauthor{\bsnm{Zhu}, \binits{B.}},
\oauthor{\bsnm{Yang}, \binits{C.}},
\oauthor{\bsnm{Dai}, \binits{J.}},
\oauthor{\bsnm{Fan}, \binits{J.}},
\oauthor{\bsnm{Qin}, \binits{Y.}},
\oauthor{\bsnm{Ye}, \binits{Y.}}:
R 2 fd 2: Fast and robust matching of multimodal remote sensing images via repeatable feature detector and rotation-invariant feature descriptor.
IEEE Transactions on Geoscience and Remote Sensing
(2023)
\end{botherref}
\endbibitem

\bibitem[\protect\citeauthoryear{Shen et~al.}{2020}]{shen20203d}
\begin{bchapter}
\bauthor{\bsnm{Shen}, \binits{W.}},
\bauthor{\bsnm{Zhang}, \binits{B.}},
\bauthor{\bsnm{Huang}, \binits{S.}},
\bauthor{\bsnm{Wei}, \binits{Z.}},
\bauthor{\bsnm{Zhang}, \binits{Q.}}:
\bctitle{3d-rotation-equivariant quaternion neural networks}.
In: \bbtitle{Computer Vision--ECCV 2020: 16th European Conference, Glasgow, UK, August 23--28, 2020, Proceedings, Part XX 16},
pp. \bfpage{531}--\blpage{547}
(\byear{2020}).
\bcomment{Springer}
\end{bchapter}
\endbibitem

\bibitem[\protect\citeauthoryear{Cohen et~al.}{2021}]{cohen2021equivariant}
\begin{botherref}
\oauthor{\bsnm{Cohen}, \binits{T.}}, et al.:
Equivariant convolutional networks.
PhD thesis,
Taco Cohen
(2021)
\end{botherref}
\endbibitem

\bibitem[\protect\citeauthoryear{Hamdi et~al.}{2022}]{Hamdi2022MVTNLM}
\begin{botherref}
\oauthor{\bsnm{Hamdi}, \binits{A.}},
\oauthor{\bsnm{AlZahrani}, \binits{F.}},
\oauthor{\bsnm{Giancola}, \binits{S.}},
\oauthor{\bsnm{Ghanem}, \binits{B.}}:
Mvtn: Learning multi-view transformations for 3d understanding.
ArXiv
\textbf{abs/2212.13462}
(2022)
\end{botherref}
\endbibitem

\bibitem[\protect\citeauthoryear{Dong et~al.}{2022}]{Dong2022ViewFoolET}
\begin{botherref}
\oauthor{\bsnm{Dong}, \binits{Y.}},
\oauthor{\bsnm{Ruan}, \binits{S.}},
\oauthor{\bsnm{Su}, \binits{H.}},
\oauthor{\bsnm{Kang}, \binits{C.}},
\oauthor{\bsnm{Wei}, \binits{X.}},
\oauthor{\bsnm{Zhu}, \binits{J.}}:
Viewfool: Evaluating the robustness of visual recognition to adversarial viewpoints.
36th Conference on Neural Information Processing Systems (NeurIPS)
\textbf{abs/2210.03895}
(2022)
\end{botherref}
\endbibitem

\bibitem[\protect\citeauthoryear{Ruan et~al.}{2023}]{Ruan2023TowardsVV}
\begin{botherref}
\oauthor{\bsnm{Ruan}, \binits{S.}},
\oauthor{\bsnm{Dong}, \binits{Y.}},
\oauthor{\bsnm{Su}, \binits{H.}},
\oauthor{\bsnm{Peng}, \binits{J.}},
\oauthor{\bsnm{Chen}, \binits{N.}},
\oauthor{\bsnm{Wei}, \binits{X.}}:
Towards viewpoint-invariant visual recognition via adversarial training.
2023 IEEE/CVF International Conference on Computer Vision (ICCV),
4686--4696
(2023)
\end{botherref}
\endbibitem

\bibitem[\protect\citeauthoryear{Milletari et~al.}{2016}]{milletari2016v}
\begin{bchapter}
\bauthor{\bsnm{Milletari}, \binits{F.}},
\bauthor{\bsnm{Navab}, \binits{N.}},
\bauthor{\bsnm{Ahmadi}, \binits{S.-A.}}:
\bctitle{V-net: Fully convolutional neural networks for volumetric medical image segmentation}.
In: \bbtitle{2016 Fourth International Conference on 3D Vision (3DV)},
pp. \bfpage{565}--\blpage{571}
(\byear{2016}).
\bcomment{Ieee}
\end{bchapter}
\endbibitem

\bibitem[\protect\citeauthoryear{Hamdi et~al.}{2022}]{hamdi2022mvtn}
\begin{botherref}
\oauthor{\bsnm{Hamdi}, \binits{A.}},
\oauthor{\bsnm{AlZahrani}, \binits{F.}},
\oauthor{\bsnm{Giancola}, \binits{S.}},
\oauthor{\bsnm{Ghanem}, \binits{B.}}:
Mvtn: Learning multi-view transformations for 3d understanding.
arXiv preprint arXiv:2212.13462
(2022)
\end{botherref}
\endbibitem

\bibitem[\protect\citeauthoryear{Mescheder et~al.}{2019}]{mescheder2019occupancy}
\begin{bchapter}
\bauthor{\bsnm{Mescheder}, \binits{L.}},
\bauthor{\bsnm{Oechsle}, \binits{M.}},
\bauthor{\bsnm{Niemeyer}, \binits{M.}},
\bauthor{\bsnm{Nowozin}, \binits{S.}},
\bauthor{\bsnm{Geiger}, \binits{A.}}:
\bctitle{Occupancy networks: Learning 3d reconstruction in function space}.
In: \bbtitle{Proceedings of the IEEE/CVF Conference on Computer Vision and Pattern Recognition},
pp. \bfpage{4460}--\blpage{4470}
(\byear{2019})
\end{bchapter}
\endbibitem

\bibitem[\protect\citeauthoryear{Sun et~al.}{2018}]{sun2018pix3d}
\begin{bchapter}
\bauthor{\bsnm{Sun}, \binits{X.}},
\bauthor{\bsnm{Wu}, \binits{J.}},
\bauthor{\bsnm{Zhang}, \binits{X.}},
\bauthor{\bsnm{Zhang}, \binits{Z.}},
\bauthor{\bsnm{Zhang}, \binits{C.}},
\bauthor{\bsnm{Xue}, \binits{T.}},
\bauthor{\bsnm{Tenenbaum}, \binits{J.B.}},
\bauthor{\bsnm{Freeman}, \binits{W.T.}}:
\bctitle{Pix3d: Dataset and methods for single-image 3d shape modeling}.
In: \bbtitle{Proceedings of the IEEE Conference on Computer Vision and Pattern Recognition},
pp. \bfpage{2974}--\blpage{2983}
(\byear{2018})
\end{bchapter}
\endbibitem

\end{thebibliography}

\end{document}